\documentclass[letterpaper, 10 pt, conference]{ieeeconf}  

\IEEEoverridecommandlockouts                              

\usepackage{graphicx} 
\usepackage{xcolor} 
\usepackage{amsmath} 
\usepackage{amssymb}  
\usepackage{multirow}
\usepackage{url}
\usepackage{colortbl}
\usepackage{tikz}

\definecolor{cgray}{rgb}{0.8, 0.8, 0.8}

\def\eg{\emph{e.g.,}~} 
\def\ie{\emph{i.e.,}~}

\title{\LARGE \bf Efficient Motion Prediction: A Lightweight \& Accurate Trajectory Prediction Model With Fast Training and Inference Speed}

\author{Alexander Prutsch$^{1}$, Horst Bischof$^{1}$ and Horst Possegger$^{1}$
\thanks{$^{1}$ Alexander Prutsch, Horst Bischof and Horst Possegger are with the Institute
of Computer Graphics and Vision, Graz University of Technology. Corresponding author: {\tt alexander.prutsch@tugraz.at}}
}

\begin{document}

\maketitle
\thispagestyle{empty}
\pagestyle{empty}

\begin{abstract}
For efficient and safe autonomous driving, it is essential that autonomous vehicles can predict the motion of other traffic agents.
While highly accurate, current motion prediction models often impose significant challenges in terms of training resource requirements and deployment on embedded hardware.
We propose a new efficient motion prediction model, which achieves highly competitive benchmark results while training only a few hours on a single GPU.
Due to our lightweight architectural choices and the focus on reducing the required training resources, our model can easily be applied to custom datasets.
Furthermore, its low inference latency makes it particularly suitable for deployment in autonomous applications with limited computing resources.
\end{abstract}

\section{INTRODUCTION}
Predicting the motion of other traffic agents is an essential task for the efficient and safe operation of self-driving robots.
This applies to a wide range of use cases, \ie autonomous robots in intralogistics and autonomous vehicles in street traffic.
Motion prediction algorithms use environment data, \eg road typologies and historical movement data of agents.
Hence, methods are usually built on top of basic perception tasks like object detection and tracking, and combine them with static information, \ie high-definition~(HD) maps.
Subsequently, autonomous systems use the predicted future trajectories of other traffic participants as input to planning modules.
This allows the autonomous system to proactively react to its environment.
As a benefit, self-driving vehicles can move more fluently in traffic and increase safety by taking the future behavior of others into account.

Extensive research on autonomous driving cars yields great results for trajectory prediction methods.
In addition, various large-scale benchmark datasets are available for evaluating motion prediction models on street traffic, \eg Argoverse~1~(AV1)~\cite{chang2019argoverse}, Argoverse~2~(AV2)~\cite{wilson2021argoverse}, nuScenes~\cite{caesar2020nuscenes} and Waymo Open Motion Dataset~(WOMD)~\cite{ettinger2021large}.
In recent years, transformer-based encoder-decoder methods, \eg\cite{liu2021multimodal, shi2022motion, nayakanti2023wayformer, cheng2023forecast, lan2023sept}, showed a performance advantage over methods that are CNN-based, \eg\cite{phan2020covernet, chai2020multipath}, graph-centric methods, \eg\cite{liang2020learning, zeng2021lanercnn}, or recurrent neural networks-based approaches, \eg\cite{varadarajan2022multipath}.

While yielding highly accurate predictions and overall excellent results, transformer-based methods are resource-demanding, especially for training the respective models as outlined \eg in \cite{shi2022motion, zhou2023query, wang2023prophnet, zhang2023hptr}.
This yields to manifold practical implications: without significant resource investments, it is impossible to train these methods on custom datasets.
Also, adapting the models to use-case-specific requirements is difficult.
Furthermore, the deployment on embedded hardware, \eg for use on mobile robots, is non-trivial.

\begin{figure}[t]
    \centering
    \includegraphics[trim={0.4cm, 0.4cm, 0.4cm, 0.2cm}, clip, width=0.75\linewidth]{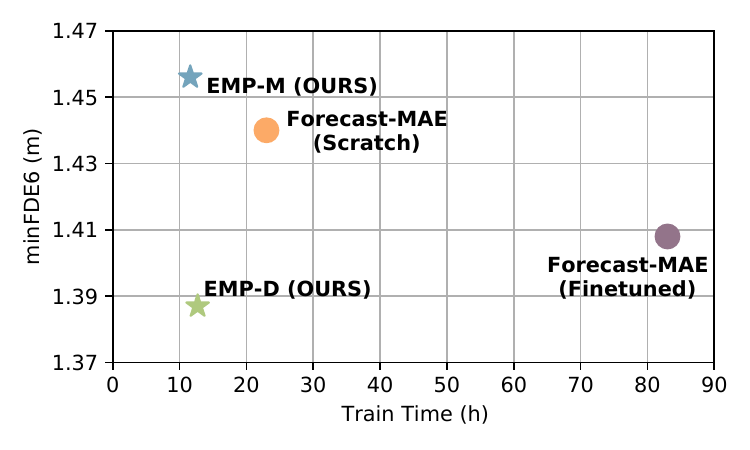}
    \includegraphics[trim={0.4cm, 0.4cm, 0.4cm, 0.2cm}, clip, width=0.75\linewidth]{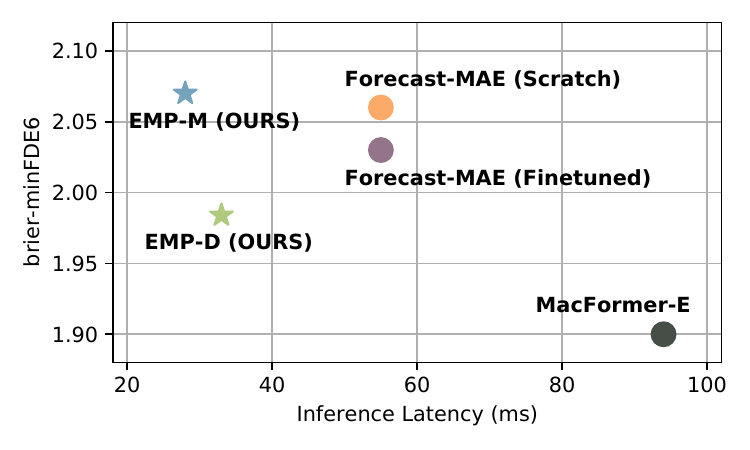}
    \vspace{-0.1cm}
    \caption{Resource and accuracy trade-off of our model compared to Forecast-MAE~\cite{cheng2023forecast} and MacFormer~\cite{feng2023macformer}. Top: Required training time on a single NVIDIA V100 GPU vs. the minFDE$_6$ on the Argoverse~2 validation set. Bottom: Inference latency for predicting 32 scenarios on a single NVIDIA RTX 2080 TI vs. the brier-minFDE$_6$ on the AV2 test set.}
    \label{fig:teaser}
    \vspace{-0.4cm}
\end{figure}

To overcome the shortcomings of resource-demanding trajectory prediction models, we propose a new \textbf{e}fficient \textbf{m}otion \textbf{p}rediction model~(EMP).
The design of EMP focuses on maximizing performance given a restricted training infrastructure.
We show that our method achieves competitive results on the challenging AV2 dataset, despite training only for a few hours on a single NVIDIA V100 GPU with 16 gigabytes VRAM.
We also highlight that our model achieves excellent inference speed on different GPU architectures.

Our model architecture builds on recent advances~\cite{cheng2023forecast, lan2023sept}, which show that excellent performance can also be achieved using only simplistic network blocks.
This contrasts with other previous works, such as \cite{shi2022motion, zhou2023query}, where improvements were mainly achieved by complex model design or explicit modeling of previous knowledge, \ie goal state candidates.

Compared to other methods that also put a focus on inference speed (\eg\cite{feng2023macformer, wang2023prophnet, feng2023macformer}), we explicitly also emphasize training speed.
Compared to Forecast-MAE (one of the few methods which can be trained on a single GPU with reasonable batch-size), we successfully speed-up training by nearly 100\% while achieving even better scores.
In addition, we perform our main evaluation on the much more complex AV2, whereas many highly efficient methods, \eg\cite{aydemir2023adapt} are only evaluated on the simpler AV1 dataset. 

Our model architecture is based on standard transformer blocks for encoding agent histories, road topology and scene information.
For decoding the future trajectories and confidence scores, we experiment with different decoder architectures.
We compare the use of a simple multi-layer perceptron-based decoder with a sophisticated transformer-based method.

In summary, our main contributions include:
\begin{itemize}
    \item We propose a new, simplistic but highly effective, transformer-based trajectory prediction model achieving results competitive to the state-of-the-art without incorporating prior knowledge or time intensive pre-training.
    \item We put an explicit emphasis on limiting the required resources for training our model, making the model easily adoptable by practitioners.
    \item We provide preliminary results on the AV1 dataset and execute a detailed model study on the complex AV2 dataset, where we report training resources, inference speed and prediction accuracy.
\end{itemize}

\section{RELATED WORK}
\begin{figure*}[t]
    \vspace{0.17cm}
    \centering
    \includegraphics[trim={0cm, 0.1cm, 0cm, 0cm}, clip, width=0.83\linewidth]{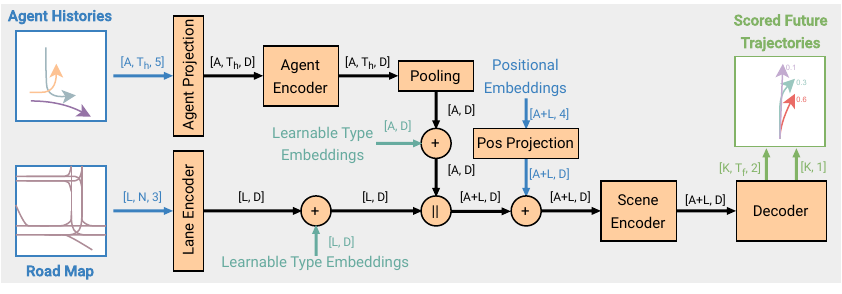}
    \vspace{-0.25cm}
    \caption{General overview of our EMP model architecture. $||$ denotes concatenation.}
    \label{fig:arch}
    \vspace{-0.4cm}
\end{figure*}

\subsection{Transformer-based Motion Prediction Models}
In recent years, the use of transformer methods for motion prediction has been widely studied.
Early works use attention mechanisms to model the interaction between different agents in traffic scenes, \eg~\cite{vemula2018social, li2020end, mercat2020multi}.
LaneGCN~\cite{liang2020learning} uses an attention-based network to fuse actor and map features, which are represented in a graph structure.
The fusion network executes actor-to-lane, actor-to-actor, lane-to-lane and lane-to-actor attention to model the relationship between the target agent, road topology and other agents.

mmTransformer~\cite{liu2021multimodal} uses a stacked transformer backbone. 
Its transformer blocks aggregate different input modalities (\ie map information, target agent history and agent interactions) using cross-attention.
They employ a custom training strategy and predict future trajectories based on fixed proposals.
Wayformer~\cite{nayakanti2023wayformer} studies different attention types and multiple strategies to fuse multi-modal input data.
Implementing a transformer-based encoder and decoder structure, they achieved state-of-the-art results in 2022 on the Argoverse~1~(AV1) and Waymo Open Motion Dataset~(WOMD) using a modality-agnostic early fusion method.

MTR~\cite{shi2022motion} introduces a trajectory decoder based on the detection transformer~(DETR)~\cite{carion2020end}.
It combines a global intent localization module with a local movement refinement by iteratively updating the predictions based on a set of motion queries.
To this end, they initially cluster the training data endpoints into a fixed set of intent points, which are then used to create the motion query.
Hence, each motion query is used to model a specific agent intent.

QCNet~\cite{zhou2023query} also utilizes a DETR-like decoder structure.
To overcome the limitations of anchor-based approaches (dependence on prior knowledge) and anchor-free approaches (mode collapse), it utilizes a two-stage decoding pipeline. 
First, trajectory proposals are generated using anchor-free, learnable queries.
Second, the proposals are used as anchors for generating the final predictions.
Both stages recurrently apply cross-attention to scene and agent tokens, as well as, self-attention across mode queries.

In recent work, Forecast-MAE~\cite{cheng2023forecast} shows that a simplistic network architecture without additional prior knowledge is also capable of yielding competitive results on the AV2 dataset.
Furthermore, they boost their prediction accuracy by implementing a pre-training scheme for masked agents and lane tracks.
Forecast-MAE utilizes a neighborhood attention~\cite{hassani2023neighborhood} to encode agent histories and a mini-PointNet~\cite{qi2017pointnet} based approach to encode lane information.
Then, they concatenate actor and lane tokens, add positional information and encode it using self-attention via standard transformer blocks.
They use a simple multi-layer perceptron~(MLP)-based decoder for generating the trajectories and scores.
SEPT~\cite{lan2023sept} also follows the idea of using a simple network structure and pre-training.
They utilize self-attention and max-pooling on agent histories to compute the agent tokens.
Following, they use a spatial encoder to learn the spatio-temporal information of agents and the road network.
As a decoder, they also propose the use of a DETR-like module that does cross-attention of learnable queries to the scene tokens.
Additionally, the pre-training tasks differ from Forecast-MAE, and the token granularity is finer (waypoints instead of whole history/future and short road vectors compared to whole segments).
Also, the bigger model size is a likely reason for the performance advantage over Forecast-MAE.
At the moment, SEPT achieves state-of-the-art results on AV1 and AV2.
A major shortcoming of these approaches, however, is that training/inference is resource intensive -- we address this with our more lightweight prediction model.

\subsection{Lightweight Motion Prediction Models}
HiVT~\cite{zhou2022hivt} hierarchically applies transformer blocks for efficient learning of spatio-temporal information.
The scene is split into different regions, which are then processed by local encoders.
Following this, a global interaction module is used to model the interaction between different regions.
They provide a detailed study on the influence of the local region size, which yields an information vs.~complexity trade-off and, furthermore, an accuracy vs.~inference speed trade-off.

ADAPT~\cite{aydemir2023adapt}, ProphNet~\cite{wang2023prophnet} and MACFormer~\cite{feng2023macformer} are three recent motion prediction methods with an emphasis on inference speed.
ADAPT utilizes a LaneGCN~\cite{liang2020learning} inspired transformer-based encoder with modality-specific relation modeling.
Following this, they use a two-stage MLP-based decoder.
Initially, they predict endpoint candidates based on the scene features.
Next, the scene features and endpoints are fed into an MLP to generate refined endpoints.
They only evaluate on the AV1 dataset and do not provide results, \eg on the challenging AV2 dataset.

ProphNet~\cite{wang2023prophnet} also utilizes a multi-stage decoding approach after encoding the agent and road inputs in an agent-centric approach using gated MLPs~\cite{liu2021pay}.
Then, they generate proposals by cross-attending learnable queries to the agent histories and anchors by self-attention of scene features.
Subsequently, the proposals and anchors are combined to predict the trajectories.
However, ProphNet requires significant training resources, \ie 16 NVIDIA V100 GPUs with a batch size of 64, whereas our model can be trained on a single V100 with batch size 96.

MacFormer~\cite{feng2023macformer} uses a multitask optimization strategy which explicitly takes map constraints into account.
The single MacFormer models show low latency and good accuracy on the AV~1 dataset. 
On the difficult AV2 dataset, they propose a model ensemble to achieve state-of-the-art results.
Compared to our approach, their ensemble method yields slightly better accuracy, but also has a much higher latency.
Furthermore, training a model ensemble opposes our goal of using only limited training resources.

HPTR~\cite{zhang2023hptr} provides a hierarchical framework based on a new attention mechanism using K-nearest neighbor attention and relative pose encoding.
In their work, they also address the required training resources.
They train for 5–10 days on 4 NVIDIA RTX 2080 Ti GPUs (batch size 12), which is significantly less training effort than, \eg Wayformer~\cite{nayakanti2023wayformer} and ProphNet~\cite{wang2023prophnet}.
Even with fewer training resources (our models require less than 15 hours to train on a single RTX 2080 Ti with batch size 64), our model exceeds the performance of HPTR on the AV2 dataset.

\section{EFFICIENT TRAJECTORY PREDICTION}
We propose a new efficient motion prediction model (EMP) based on standard transformer blocks.
Our model follows recent ideas to use a simplistic architecture and avoid the introduction of inductive bias via prior knowledge~{\cite{cheng2023forecast, lan2023sept}.
Figure~\ref{fig:arch} gives an overview of our method architecture.

Our main focus is to build a model without processing bottlenecks, which allows fast and efficient training of the model, as well as rapid inference speed.
To this end, we also disregard the idea of pre-training proposed by~\cite{cheng2023forecast, lan2023sept}.
While pre-training allows to achieve performance gains compared to training from scratch, it heavily contradicts with our emphasis on low training resource consumption.
For reference, pre-training Forecast-MAE on the AV2 dataset would take around 60 hours on a single V100 GPU.

\subsection{Agent Encoding}
We use the 2D position of agents, agent velocity magnitude, a step counter and a mask flag (indicates if agent was observed at a given timestamp) to encode the state of each agent $a$ at time $t$ into a 5-element tensor.
The step counter is used to preserve the state sequence order during attention operations.
The agent positions are normalized with respect to the center pose of each track.
Next, we project the state tensor $\mathbb{R}^{A \times T_h \times 5}$, where $A$ corresponds to the number of agents and $T_h$ corresponds to the number of historic time steps, to our embedding space $\mathbb{R}^{A \times T_h \times D}$.
To this end, we utilize a simple linear layer.

Subsequently, we apply self-attention along the temporal dimension using standard multi-head attention transformer blocks to efficiently learn information about each agent track~\cite{lan2023sept}.
Our experiments show that this architecture yields significant speed advantages compared to other agent encoder structures, \eg\cite{cheng2023forecast}.
Then, we reduce our tokens to $\mathbb{R}^{A \times D}$ via max pooling along the temporal dimension.
Finally, we add learnable type embeddings $\mathbb{R}^{A \times D}$ to get our our final agent tokens.
The type embeddings are used to learn categorical information about the input agents.

\subsection{Lane Encoding}
To efficiently process lane information, we encode the data of each lane segment to a lane token.
Each lane segment is initially represented by a set of $N$ points in 2D space and a masking flag ($\mathbb{R}^{L \times N \times 3}$).
The points describe the shape of each lane center line w.r.t.~to the origin of each lane object.
We apply a mini PointNet-like~\cite{qi2017pointnet} lane encoder, as proposed by~\cite{cheng2023forecast}, to get $\mathbb{R}^{L \times D}$ lane embeddings.
Analog to agent encoding, we also add learnable type embeddings to get our final lane tokens.

\subsection{Scene Encoding}
We concatenate the agent tokens ($\mathbb{R}^{A \times D}$) and lane tokens ($\mathbb{R}^{L \times D}$) to form scene tokens ($\mathbb{R}^{A+L \times D}$).
Positional embeddings are added to the scene tokens to describe the spatial relationships between different tokens in the scene.
To compute the positional encoding, we project the centers of all tracks and lanes ($\mathbb{R}^{A+L \times 4}$) to our embedding space $\mathbb{R}^{A+L \times D}$ using two linear layers (GeLU activation function on the first and $D$ units each).
We define the center using $[x, y, \cos\alpha, \sin\alpha]$ where $\alpha$ is the orientation of the corresponding object.
Subsequently, we encode the scene tokens using self-attention blocks (scene encoder).

\begin{figure}[t]
    \vspace{0.08cm}
    \centering
    \includegraphics[trim={0cm, 0.3cm, 0cm, 0cm}, clip, width=0.83\linewidth]{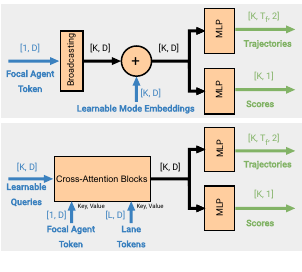}
    \vspace{-0.2cm}
    \caption{Decode module architectures: simple MLP-based decoder (for EMP-M, top) and DETR-like decoder (for EMP-D, bottom).}
    \label{fig:dec}
    \vspace{-0.4cm}
\end{figure}

\subsection{Decoding}
We experiment with two decoder architectures, shown in Figure~\ref{fig:dec}.
First, we use a pure multi-layer perception~(MLP)-based approach to achieve a very lightweight model (EMP-M).
Second, we utilize a more sophisticated transformer-based decoder (EMP-D), similar to \cite{shi2022motion, nayakanti2023wayformer, zhou2023query, lan2023sept}, who also use DETR-like decoders for trajectory prediction.

\subsubsection{MLP-based Decoder (EMP-M)}
We use learnable mode embeddings $\mathbb{R}^{K \times D}$ to efficiently generate $K$ multi-modal output trajectories for each focal agent token $\mathbb{R}^{1 \times D}$.
First, we broadcast the token to $\mathbb{R}^{6 \times D}$, then we add our learnable embeddings.
Afterwards, we use two MLPs to predict the multi-modal trajectory positions and confidence scores.
Both MLPs have a single hidden layer with $2\cdot D$ units and \textit{ReLU} activation function.
For predicting the trajectories we use a linear output layer with $2\cdot T_f$ units, where $T_f$ corresponds to the number of future time steps.
For the confidence scores we use a linear output layer with $K$ units.

\subsubsection{DETR-like Decoder (EMP-D)}
To model multi-modality, we adopt learnable query embeddings, \ie~\cite{shi2022motion, lan2023sept}.
We initialize queries $\mathbb{R}^{K \times D}$ using $K$ learnable embeddings.
Then, we process the queries using multiple cross-attention blocks.
In each block, we first perform cross-attention on the focal agent embeddings.
Second, we execute cross-attention to the lane tokens to improve alignment with the road network.
We do not apply additional cross-attention to other agent tokens~\cite{shi2022motion, zhou2023query} or mode self-attention~\cite{zhou2023query}, thus accepting a marginally lower accuracy in favor of a lightweight architecture.
Lastly, we use two MLPs to map the updated queries to the trajectories and scores. 
The MLPs have the same structure as in our pure MLP-based decoder.

\section{EXPERIMENTAL SETUP}
\begin{table*}[t]
\setlength{\tabcolsep}{4pt}
    \vspace{0.17cm}
    \caption{Results on the Argoverse~2 single agent forecasting challenge sorted in descending order by brier-minFDE$_\mathbf{6}$.}
    \vspace{-0.4cm}
    \label{tab:av2results}
    \begin{center}
    \begin{tabular}{c||ccc|cccccc}
    \hline
           & \multicolumn{3}{c|}{Validation Set} & \multicolumn{6}{c}{Test Set} \\
    Method & MR$_6$ & minADE$_6$ & minFDE$_6$ & MR$_6$ & minADE$_1$ & minFDE$_1$ & minADE$_6$ & minFDE$_6$ & brier-minFDE$_\mathbf{6}$ \\ \hline 
    FRM~\cite{park2023leveraging}                   & - & - & - & 0.29 & 2.37 & 5.93 & 0.89 & 1.81 & 2.47\\
    THOMAS~\cite{gilles2022thomas}                  & - & - & - & 0.20 & 1.95 & 4.71 & 0.88 & 1.51 & 2.16 \\
    \rowcolor{cgray}EMP-M (Ours)                    & 0.19 & 0.73 & 1.46 & 0.19 & 1.80 & 4.53 & 0.72 & 1.43 & 2.07 \\ 
    Forecast-MAE Scratch~\cite{cheng2023forecast}   & 0.19 & 0.81 & 1.44 & 0.19 & 1.85 & 4.60 & 0.73 & 1.43 & 2.06 \\
    SIMPL~\cite{zhang2024simpl}                     & - & - & - & 0.19 & 2.03 & 5.50 & 0.72 & 1.43 & 2.05 \\
    HPTR~\cite{zhang2023hptr}                       & - & - & - & 0.19 & 1.84 & 4.61 & 0.73 & 1.43 & 2.03\\
    BANet (Singe Model)~\cite{zhang2022banet}       & - & - & - & 0.18 & 1.84 & 4.70 & 0.73 & 1.39 & 2.03 \\
    Forecast-MAE Finetuned~\cite{cheng2023forecast} & 0.18 & 0.80 & 1.41 & 0.17 & 1.74 & 4.36 & 0.71 & 1.39 & 2.03 \\
    GoRela~\cite{cui2023gorela}                     & - & - & - & 0.22 & 1.82 & 4.62 & 0.76 & 1.48 & 2.01 \\
    HeteroGCN (Single Model)~\cite{gao2023dynamic}  & - & - & - & 0.18 & 1.79 & 4.53 & 0.73 & 1.37 & 2.00 \\
    MTR~\cite{shi2022motion}                        & - & - & - & \underline{0.15} & 1.74 & 4.39 & 0.73 & 1.44 & 1.98 \\
    \rowcolor{cgray}EMP-D (Ours)                    & 0.18 & 0.71 & 1.39 & 0.17 & 1.75 & 4.35 & 0.71 & 1.37 & 1.98  \\ 
    QCNet (Single Model)~\cite{zhou2023query}       &  0.16 & 0.73 & 1.27 & 0.16 & \underline{1.69} & \underline{4.30} & \underline{0.65} & \underline{1.29} & 1.91 \\ 
    MacFormer-E (Ensemble)~\cite{feng2023macformer} & - & - & - & 0.19 & - & - & 0.70 & 1.38 & 1.90 \\
    ProphNet~\cite{wang2023prophnet}                & - & - & - & 0.18 & 1.80 & 4.74 & 0.68 & 1.33 & \underline{1.88} \\
    SEPT~\cite{lan2023sept}                         & - & - & - & \textbf{0.14} & \textbf{1.49} & \textbf{3.70} & \textbf{0.61} & \textbf{1.15} & \textbf{1.74} \\
    \hline
    \end{tabular}
    \end{center}
    \vspace{-0.5cm}
\end{table*}

\subsection{Dataset}
We conduct extensive evaluations on the motion forecasting datasets Argoverse~1~(AV1)~\cite{chang2019argoverse} and Argoverse~2~(AV2)~\cite{wilson2021argoverse}.
Both dataset include map data represented as centerline-based polylines and agent trajectory data.
The sampling rate for the trajectory data is 10~Hz and for each scene a focal agent is defined.
AV1 contains short sequences (5 seconds overall) which are rather simple (most vehicles are going straight-forward).
The first 2~seconds are considered as historic context and the goal is to predict the following 3 seconds of the focal agent.
AV2 focuses on long-term prediction: sequences are 11~seconds long, where 5 seconds are context and 6~seconds should be predicted. 
The AV2 dataset consists of sequences recorded in six American cities.
Overall, it includes 250,000 non-overlapping scenarios, which corresponds to 763 hours of data.
The dataset is randomly split into a train (199,908 sequences), validation~(24,988 sequences) and test~(24,984 sequences) set.
AV1 is only collected in two cities, but has a slightly larger number of sequences (324,557 overall).

We follow the standard pre-processing setups:
For AV1, we set the radius of interest to 65~meters around the focal agent, following~\cite{liu2021multimodal}.
For AV2, we follow Forecast-MAE~\cite{cheng2023forecast}, which uses a radius of 150~meters for data aggregation.

\subsection{Metrics}
We utilize the standard evaluation metrics of the AV~2 benchmark: miss rate~(MR$_K$), minimum average displacement error~(minADE$_K$), minimum final displacement error~(minFDE$_K$) and brier minimum final displacement error (brier-minFDE$_K$).
Each metric is computed based on a set of $K$ trajectories predicted by the model.
For $k>1$ the best fitting trajectory (minimum $L^2$ distance) is used for computing the errors.
The miss rate denotes the percentage of scenarios where no predicted trajectory endpoint is within a threshold distance to the ground truth endpoint ($2~m$ for the Argoverse benchmarks).
The brier-minFDE$_K$ adds the penalty term $(1-p)^2$ to minFDE$_K$, where $p$ is the confidence score of the best matching trajectory.
Following the AV~2 benchmark, we report minADE$_1$, minFDE$_1$, minADE$_6$, minFDE$_6$, MR$_6$ and brier-minFDE$_6$.

\subsection{Implementation Details}
In our model, we utilize an embedding size $D=128$.
We stack four transformer blocks for our agent and scene encoders and use three attention blocks in our query decoder. 
In each transformer block, we use eight heads in multi-head-attention and apply normalization before the attention operation and before our feed-forward network pass.

We train our model on a single NVIDIA V100 GPU with 16~GB VRAM for 60 epochs using a AdamW~\cite{loshchilov2017decoupled} optimizer.
We utilize a batch size of 96, which is the largest feasible value given our GPU. 
For the learning rate, we use 10 warm-up epochs to increase the learning rate to 0.001, before decreasing it to 0.0001 using a cosine schedule.
We apply norm-based gradient clipping and execute weight decay.
No augmentations are used in training.
As loss function, we sum a regression loss (Huber loss~\cite{huber1964robust}) and a classification loss (cross-entropy loss).
We compute the regression loss only between the ground truth trajectory and the best fitting trajectory (smallest average displacement error). 
Additionally, we utilize an auxiliary loss, where we predict a single trajectory for each agent in the scene~\cite{cheng2023forecast}.
We apply a linear layer $[D, 2\cdot T_f]$ to predict a single trajectory from each agent token and compute the Huber loss.

Additionally, we provide further timing analysis for training on a NVIDIA RTX 4090.
We also compare inference latency on the AV2 dataset using NVIDIA RTX 2080 Ti and RTX 4090 GPUs.
Hence, we can highlight our model effectiveness on different GPU architectures and provide a broader comparison to reference methods.

\section{RESULTS AND DISCUSSION}
\begin{figure*}[t]
    \vspace{0.17cm}
    \centering
    \includegraphics[trim={6cm, 8cm, 6cm, 6cm}, clip, width=0.25\linewidth]{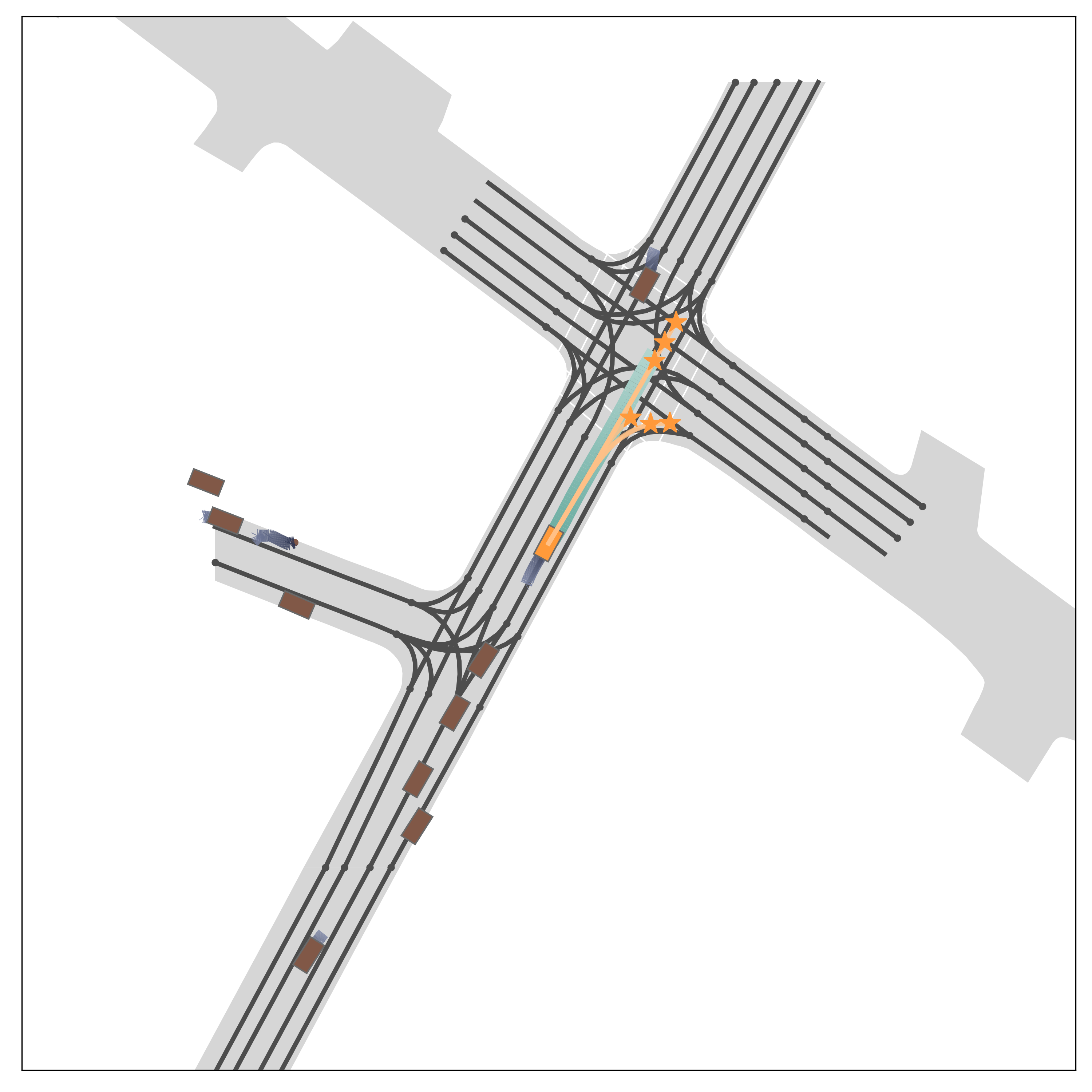}
    \hspace{0.3cm}
    \includegraphics[trim={6cm, 6cm, 6cm, 8cm}, clip, width=0.25\linewidth]{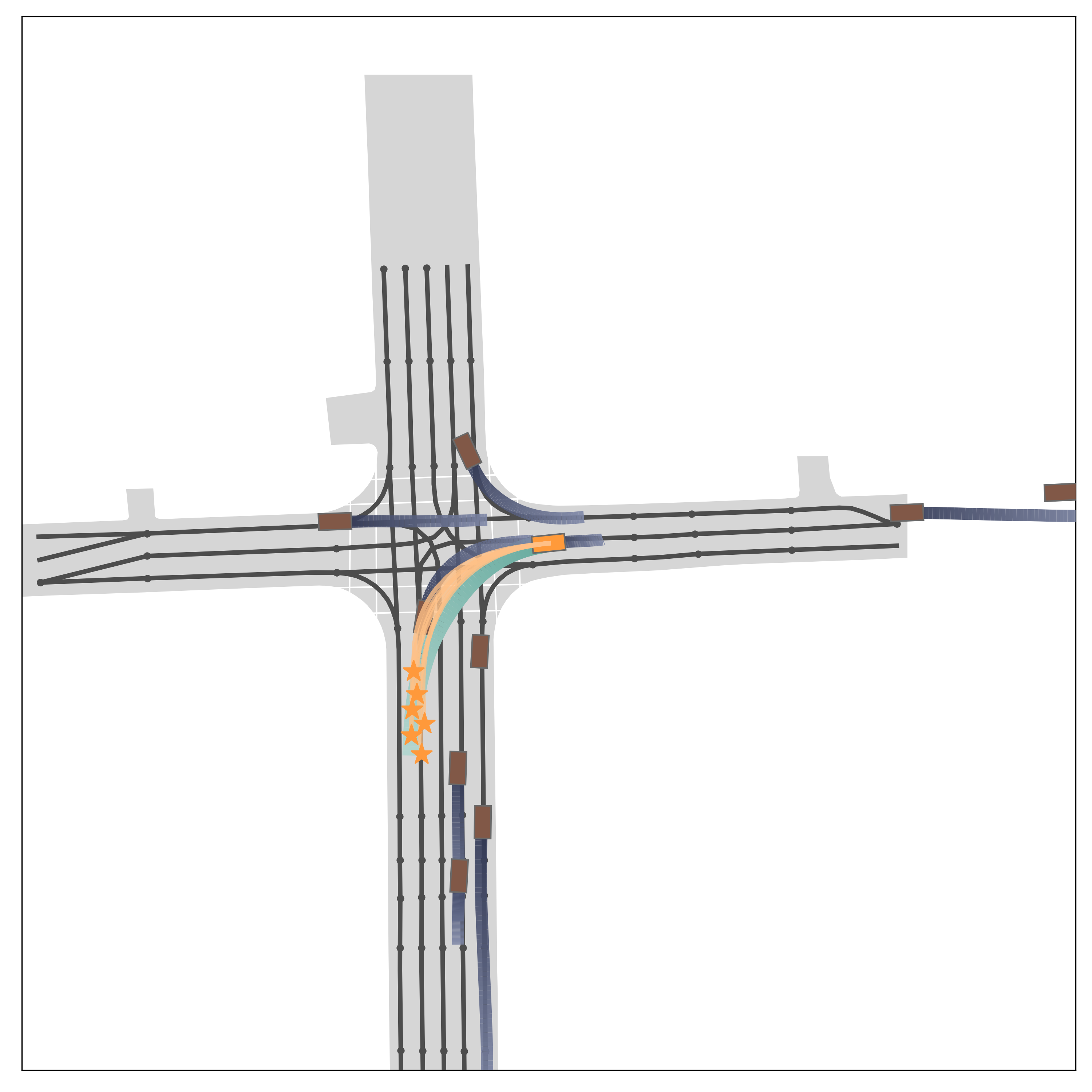}
    \hspace{0.3cm}
    \includegraphics[trim={6cm, 8cm, 6cm, 6cm}, clip, width=0.25\linewidth]{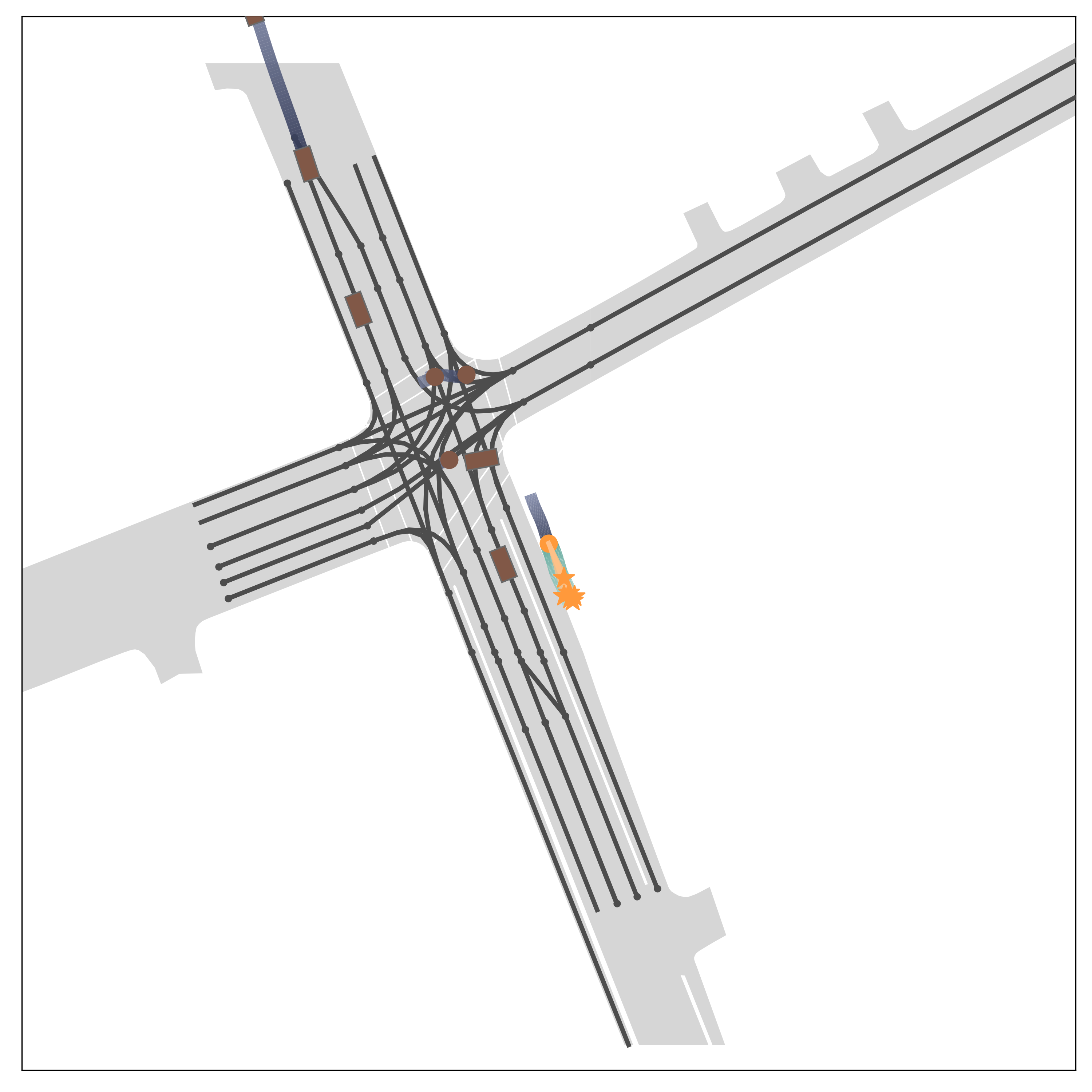}
    \vspace{-0.15cm}
    \caption{
    Exemplary EMP-D results on AV2 for 2 vehicles (left, middle) and 1 pedestrian (right), showing the focal agent (orange), predictions (orange), ground truth (turquoise), history (blue), other agents (brown) and lane centerlines (black).}
    \label{fig:res}
    \vspace{-0.25cm}
\end{figure*}

We present detailed evaluations on the challenging AV2 single-agent prediction benchmark and the AV1 benchmark.
We report results from official publications and the leaderboard\footnote{\url{https://eval.ai/web/challenges/challenge-page/1719/leaderboard/4098}}.
To study the accuracy/resource trade-off, we compare the training times and inference latency based on own experiments and reported results.

\subsection{Evaluation on AV2} 
Table~\ref{tab:av2results} summarizes the results on the AV2 single-agent prediction task. 
Due to our focus on efficiency, we report scores for single models rather than ensemble results when available.
Our simple model (EMP-M) using a MLP-based decoder achieves decent results on the validation and test set.
It yields a brier-minFDE$_6$, which is significantly lower than, \eg~THOMAS~\cite{gilles2022thomas}, and only slightly higher than Forecast-MAE~\cite{cheng2023forecast} (without pre-training).

Using a DETR-like decoder (EMP-D), we get competitive results on the AV2 test set.
EMP-D achieves a brier-minFDE$_6$ of 1.98, which is the same as the sophisticated MTR~\cite{shi2022motion} and better than recent methods like GoRela~\cite{cui2023gorela} and SIMPL~\cite{zhang2024simpl}.
In comparison to Forecast-MAE,~\cite{cheng2023forecast} which also shares the idea of a compact architecture, our model improves the brier-minFDE$_6$ by 0.08 (compared to Forecast-MAE without pre-training) and by 0.05 (finetuned Forecast-MAE).

A few models, \ie QCNet~\cite{zhou2023query}, MacFormer-E~\cite{feng2023macformer}, ProphNet~\cite{wang2023prophnet} and SEPT~\cite{lan2023sept} outperform our approach at the expense of higher training resources and slower inference.
At the time of submission, SEPT~\cite{lan2023sept} ranks third on the public leaderboard, whereas the top methods have not yet been published.
In Figure \ref{fig:res}, we show some prediction results of our EMP-D model on the AV2 dataset.

\subsection{Comparison of Training and Inference Resources}
Following, we compare the required training resources and inference latency of the methods listed in Table~\ref{tab:av2results}.
For experiments on QCNet\footnote{\url{https://github.com/ZikangZhou/QCNet}} and Forecast-MAE\footnote{\url{https://github.com/jchengai/forecast-mae}} we utilize the official code implementations.
To date, no code implementations have been provided for the high accurate methods MacFormer, ProphNet and SEPT.

The required GPU infrastructure and training duration listed in Table~\ref{tab:trainspeed} show that our model requires very small resources.
In terms of the prediction accuracy, our EMP even outperforms several state-of-the-art models which require substantially more training resources.
The more complex DETR-like decoder variant, EMP-D, only requires slightly more resources than EMP-M.
This shows that a sophisticated decoder design enables much better scores while only slightly increasing resource demands.

Generally, better performing models utilize significantly larger training infrastructure. 
Only SEPT states that it also supports training on a single RTX 3090 (which is more powerful and has more VRAM than our V100). 
But SEPT has three times the parameters as our model and its two phase training process indicates a much longer training process overall.
Unfortunately, SEPT does not report training times and no code implementation is available at the moment.

We experimentally compare the train speed of our models to Forecast-MAE, which is a recent method also supporting training on a single GPU.
Our EMP-D model requires only nearly half of the training time, but still outperforms the Forecast-MAE model from scratch by a large margin.
Additionally, our EMP-D model even exceeds the performance of the pre-trained Forecast-MAE.
Forecast-MAE would require over 60 hour for pretraining on our GPU.

\begin{table}[t]
\setlength{\tabcolsep}{3pt}
    \caption{Overview of model size and required training infrastructure for methods listed in Table~\ref{tab:av2results}. GPU column indicates type of NVIDIA GPUs used for training. Training times are reported on the AV2 dataset. * denotes measurements done by us.}
    \vspace{-0.4cm}
    \label{tab:trainspeed}
    \begin{center}
    \begin{tabular}{c||ccrl}
    \hline
    Method                                     & Params                & Training Time & \multicolumn{2}{c}{GPUs} \\ \hline \hline 
    ProphNet~\cite{wang2023prophnet}               & -                     & -             & 16x & V100 \\ \hline
    GoRela~\cite{cui2023gorela}                    & -                     & -             & 16x & T4\\ \hline
    QCNet~\cite{zhou2023query}                     & 7.7M                  & -             & 8x  & RTX 3090 \\ \hline
    SIMPL~\cite{zhang2024simpl}                    & 1.9M                  & -             & 8x  & RTX 3090 \\ \hline
    MTR~\cite{shi2022motion}                       & -                     & -             & 8x  & RTX 8000 \\ \hline
    MacFormer-E~\cite{feng2023macformer}           & 5x 2.5M               & -             & 8x  & RTX 2080 Ti \\ \hline
    HPTR~\cite{zhang2023hptr}                      & $\sim$15M             & 5-10 days     & 4x  & RTX 2080 Ti\\ \hline
    SEPT~\cite{lan2023sept}                        & 9.6M                  & -             & 1x  & RTX 3090 Ti \\ \hline
    THOMAS~\cite{gilles2022thomas}                 & -                     & -             & 1x  & RTX 2080 \\ \hline       
    Forecast-MAE~\cite{cheng2023forecast}          & \multirow{2}{*}{1.9M} & $>80$h*       & 1x  & V100 \\ 
    (Finetuned)                                    &                       & $>40$h*       & 1x  & RTX 4090 \\ \hline
    Forecast-MAE~\cite{cheng2023forecast}          & \multirow{2}{*}{1.9M} & $\sim22.7$h*  & 1x  & V100 \\ 
    (Scratch - w/o pre-train)                      &                       & $\sim11.5$h*  & 1x  & RTX 4090 \\ \hline
    \rowcolor{cgray}                               &                       & $\sim12.7$h*  & 1x  & V100 \\
    \rowcolor{cgray}\multirow{-2}{*}{EMP-D (Ours)} & \multirow{-2}{*}{3.2M}& $\sim6.3$h*   & 1x  & RTX 4090 \\ \hline
    \rowcolor{cgray}                               &                       & $\sim11.6$h*  & 1x  & V100 \\
    \rowcolor{cgray}\multirow{-2}{*}{EMP-M (Ours)} & \multirow{-2}{*}{2.0M}& $\sim6$h*     & 1x  & RTX 4090 \\ \hline
    
    \end{tabular}
    \end{center}
    \vspace{-0.6cm}
\end{table}

\begin{table}[t]
\setlength{\tabcolsep}{4pt}
    \caption{Inference speed on AV2 (latency for batch predictions of 32 scenarios). * denotes measurements done by us.}
    \vspace{-0.4cm}
    \label{tab:infspeed}
    \begin{center}
    \begin{tabular}{c|cc}
    \hline
    GPU                           & Method                                          & Latency \\ \hline 
                                  & QCNet~\cite{zhou2023query}                      & 679 ms* \\
                                  & Forecast-MAE~\cite{cheng2023forecast}           & 69 ms* \\
                                  & \cellcolor{cgray}EMP-D (Ours)                   & \cellcolor{cgray}\underline{37 ms*} \\ 
    \multirow{-4}{*}{V100}        & \cellcolor{cgray}EMP-M (Ours)                   & \cellcolor{cgray}\textbf{28 ms*} \\  \hline
                                  & MacFormer-E (ensemble)~\cite{feng2023macformer} & \phantom{ }94 ms\phantom{*} \\
                                  & Forecast-MAE~\cite{cheng2023forecast}           & \phantom{ }55 ms* \\
                                  & \cellcolor{cgray}EMP-D (Ours)                   & \cellcolor{cgray}\phantom{ }\underline{33 ms*} \\ 
    \multirow{-4}{*}{RTX 2080 Ti} & \cellcolor{cgray}EMP-M (Ours)                   & \cellcolor{cgray}\phantom{ }\textbf{30 ms*} \\ \hline
                                  & QCNet~\cite{zhou2023query}                      & 318 ms* \\
                                  & Forecast-MAE~\cite{cheng2023forecast}           & \phantom{ }23 ms* \\
                                  & \cellcolor{cgray}EMP-D (Ours)                   & \cellcolor{cgray}\phantom{ }\underline{13 ms*} \\ 
    \multirow{-4}{*}{RTX 4090}    & \cellcolor{cgray}EMP-M (Ours)                   & \cellcolor{cgray}\phantom{ }\textbf{11 ms*} \\ 
    \hline
    \end{tabular}
    \end{center}
    
\setlength{\tabcolsep}{3pt}
    \caption{Preliminary results on the Argoverse~1 validation split.}
    \vspace{-0.4cm}
    \label{tab:av1results}
    \begin{center}
    \begin{tabular}{c||ccc}
    \hline
                                                   & \multicolumn{3}{c}{Validation Set} \\
    Method                                         & MR$_6$ & minADE$_6$       & minFDE$_6$ \\ \hline 
    Forecast-MAE Scratch~\cite{cheng2023forecast}  & 0.103  & 0.74             & 1.10 \\ 
    Forecast-MAE Finetuned~\cite{cheng2023forecast}& 0.095  & 0.71             & 1.05  \\ 
    \rowcolor{cgray}EMP-M (Ours)                   & 0.096  & \underline{0.63} & \underline{1.04} \\ 
    \rowcolor{cgray}EMP-D (Ours)                   & \textbf{0.090} & \textbf{0.63}    & \textbf{1.02} \\ 
    \hline
    \end{tabular}
    \end{center} 
    \vspace{-0.55cm}
\end{table}

Table~\ref{tab:infspeed} shows the inference speed of different methods on the AV2 dataset.
We measure the latency for predicting a batch of 32 scenarios using three different GPU types: NVIDIA V100, NVIDIA RTX 2080 Ti and NVIDIA RTX 4090.
On all GPUs our EMP-M ranks first and EMP-D second. 
Both models are almost twice as fast as Forecast-MAE.
On a RTX 2080 Ti our models require less than 35~ms whereas a latency of 94~ms is reported for MacFormer-E~\cite{feng2023macformer}.
Our approach is more than 20 times faster than QCNet~\cite{zhou2023query} on V100 and RTX 4090 GPUs.
SEPT~\cite{lan2023sept} reports that its inference is twice as fast as QCNet, from which we can derive that our model is substantially faster than SEPT.
ProphNet achieves an inference latency of 27.4~ms for predicting one scenario on a V100 GPU, but no measurements for multi-scenario settings have been provided.
HeteroGCN~\cite{gao2023dynamic} reports an inference time of 57.03~ms for batch size 1 on an RTX 2080, which also indicates a much higher latency than our model.

\subsection{Evaluation on AV1}
Finally, we present preliminary results on the AV1 dataset in Table~\ref{tab:av1results}.
Achieving an optimal latency/accuracy trade-off requires adapting the data collection to the scenario complexity (AV1 sequences are shorter and most vehicles go straight).
As we focus on AV2, which better reflects the challenges of real-world driving, we did not optimize the preprocessing or hyperparameters for AV1.
Our models still outperform the most closely related Forecast-MAE despite needing less training and inference resources.

\addtolength{\textheight}{-0.8cm}   

\section{CONCLUSIONS}
We propose a new \textbf{e}fficient \textbf{m}otion \textbf{p}rediction (EMP) architecture based on standard transformer blocks.
Our approach yields highly competitive results on the challenging AV2 dataset without incorporating prior-knowledge or using pre-training.
Even with restricted training time our effective model architecture outperforms recent architectures requiring vast resources.
Additionally, we highlight the efficiency of our model by comparing inference latency on different GPU architectures.
As a result, our model allows easier finetuning on new data and is also suitable for deployment to autonomous vehicles with restricted compute resources.
However, a few models with significantly higher resource demands achieve better accuracy than ours.
Therefore, it is important to consider the balance between required accuracy and feasibility of a model for each specific use case.






\bibliography{template/IEEEabrv, chapters/ref}

\end{document}